**BOA for Nurse Scheduling**

Jingpeng Li and Uwe Aickelin

{jpl, uxa}cs.nott.ac.uk, School of Computer Science and IT, University of Nottingham, Nottingham, NG8 1BB, United Kingdom

Abstract: Our research has shown that schedules can be built mimicking a human scheduler by using a set of rules that involve domain knowledge. This chapter presents a Bayesian Optimization Algorithm (BOA) for the nurse scheduling problem that chooses such suitable scheduling rules from a set for each nurse's assignment. Based on the idea of using probabilistic models, the BOA builds a Bayesian network for the set of promising solutions and samples these networks to generate new candidate solutions. Computational results from 52 real data instances demonstrate the success of this approach. It is also suggested that the learning mechanism in the proposed algorithm may be suitable for other scheduling problems.

Keywords: Nurse Scheduling, Bayesian Optimization Algorithm, Probabilistic Modeling.

## 1 Introduction

Personnel scheduling problems have been studied extensively over the past three decades (see survey papers by Baker 1976; Tien and Kamiyama 1982; Bradley and Martin 1990, Bechtold et al. 1991; Ernst et al., 2004). Personnel scheduling is the problem of assigning employees to shifts or duties over a scheduling period so that certain constraints (organizational and personal) are satisfied. It involves constructing a schedule for each employee within an organization in order for a set of tasks to be fulfilled. In the domain of healthcare, this is particularly challenging because of the presence of a range of different staff requirements on different days and shifts. In addition, unlike many other organizations, healthcare institutions work twenty-four hours a day for every single day of the year. In this chapter, we focus on the development of new approaches for nurse scheduling.

Most nurse scheduling problems are extremely difficult and complex. Tien and Kamiyama (1982), for example, say nurse scheduling is more complex than the traveling salesman problem. A general overview of various approaches for nurse scheduling can be found in Sitompul and Randhawa (1990), Cheang et al. (2003) and Burke et al. (2004). Early research (Warner and Prawda 1972; Miller et al. 1976; Trivedi and Warner 1976) concentrated on the development of mathematical programming models. To reduce computational complexity, researchers had to restrict the problem dimensions and consider a smaller size of constraints in their models, resulting in solutions that are too simple to be applied in real hospital situations.

The above observations have led to other attempts, trying to solve the real nurse scheduling problems within reasonable time. Besides heuristic approaches (e.g., Smith et al. 1979; Blau and Sear 1983), artifi-cial intelligence approaches such as constraint programming (Okada and Okada 1988), expert systems (Chen and Yeung 1993) and knowledge based systems (Scott and Simpson 1998) were investigated with some success. Since the 1990's, most papers have tackled the problem with metaheuristic approaches, including simulated annealing (Isken and Hancock 1991), tabu search (Dowsland 1998) and genetic algorithms (Aickelin and Dowsland 2003).

This chapter solves a nurse scheduling problem arising at a major UK hospital (Aickelin and Dowsland 2000; Dowsland and Thompson 2000). In this problem, the number of nurses is fixed (up to 30) and the target is to create a weekly schedule by assigning each nurse one out of up to 411 predefined shift patterns in the most efficient way. Due to the limitation of the traditional mathematical programming approach, a number of meta-heuristic approaches have been explored for this problem. For example, in (Aickelin and Dowsland 2000, 2003; Aickelin and White 2004) various approaches based on Genetic Algorithms (GAs) are presented, in (Li and Aickelin 2004) an approach based on a learning classifier system is investigated, in (Burke et al. 2003) a tabu search hyperheuristic is introduced, and in (Li and Aickelin 2003) a Bayesian Optimization Algorithm (BOA) is described. In this chapter, we demonstrate a novel BOA approach.

In our proposed BOA, we try to solve the problem by applying a suitable rule, from a set containing a number of available rules, for each nurse's assignment. A schedule can be then created from a rule string (or a rule sequence) corresponding to nurses from the first to the last. To evolve the rule strings Bayesian networks (Pearl 1988) are used. The Bayesian network in our case is a directed acyclic graph with each node corresponding to a nurse/rule pair, by which a schedule will be constructed step by step. The causal relationship between two nodes is represented by a directed edge between the two nodes.

Based on the idea of using probabilistic models, the BOA builds a Bayesian network for the set of promising solutions and generates new candidate solutions by sampling these networks. (Pelikan et al. 1999; Pelikan 2005). In our proposed BOA for nurse scheduling, the conditional probabilities of all nurse/rule pairs in the network are first computed according to an initial set of rule strings. Subsequently, new instances of each node are generated by using conditional probabilities to obtain new rule strings. A new set of rule strings is thus generated, some of which will replace previous strings based on roulette-wheel fitness selection. If the stopping criterion is not reached, the conditional probabilities for all nodes in the network are updated again using the current set of rule strings.

## 2 The Nurse Scheduling Problem

The problem described in this chapter is that of creating weekly schedules for wards containing up to 30 nurses at a major UK hospital. These schedules must respect working contracts and meet the demand for a given number of nurses of different grades on each shift, while being perceived to be fair by the nurse themselves. The day is partitioned into three shifts: two day shifts called 'earlies' and 'lates', and a longer night shift. Until the final scheduling stage, 'earlies' and 'lates' are merged into day shifts. Due to hospital policy, a nurse would normally work either days or nights in a given week, and due to the difference in shift length, a full week's work normally includes more days than nights. For example, a full-time nurse works 5 days or 4 nights, whereas a part-time nurse works 4 days or 3 nights, 3 days or 3 nights, or 3 days or 2 nights. However, exceptions are possible and some nurses specifically must work both day- and night-shifts in one week.

As described in (Dowsland and Thompson 2000), the problem can be decomposed into three independent stages. The first stage uses a knapsack model to check if there are enough nurses to meet demand. If not, additional nurses are allocated to the ward, so that the second stage will always admit a feasible solution. The second stage is the most difficult and is concerned with the actual allocations of days or nights to be worked by each nurse. The third stage then uses a network flow model to assign those on days to 'earlies' and 'lates'. The first and the third stages, and in this chapter we only concern with the highly constrained second stage.

The numbers of days or nights to be worked by each nurse defines the set of feasible weekly work patterns for that nurse. These will be referred to as shift patterns in the following. For example, (1111100 0000000) would be a pattern where the nurse works the first 5 days and no nights. Depending on the working hours of a nurse, there are a limited number of shift patterns available to her/him. Typically, there will be between 20 and 30 nurses per ward, 3 grade-bands, 9 part time options and 411 different shift patterns. Depending on the nurses' preferences, the recent history of patterns worked, and the overall attractiveness of the pattern, a penalty cost is allocated to each nurse-shift pattern pair. These values were set in close consultation with the hospital and range from 0 (perfect) to 100 (unacceptable), with a bias to lower values. Further details can be found in (Dowsland 1998).

This second stage problem is described as follows. Each possible weekly shift pattern for a given nurse can be represented as a 0-1 vector with 14 elements, where the first 7 elements represent the 7 days of the week and the last 7 elements the corresponding 7 nights of the week. A '1'/'0' in the vector denotes a scheduled day or night 'on'/'off'. For each nurse there are a limited number of shift patterns available to her/him. For instance, a full-time nurse working either 5 days or 4 nights has a total of 21 (i.e. $C_7^5$) feasible day shift patterns and 35 (i.e. $C_7^4$) feasible night shift patterns. Typically, there are between 20 and 30 nurses per ward, 3 grade-bands, 9 part time options and 411 different shift patterns. Depending on the nurses' preferences, the recent history of patterns worked, and the overall attractiveness of the pattern, a penalty cost is allocated to each nurse-shift pattern pair. These values were set in close consultation with the hospital and range from 0 (perfect) to 100 (unacceptable), with a bias to lower values. Further details can be found in (Dowsland 1998).

This problem can be formulated as follows. The decision variable $x_{ij}$ assumes 1 if nurse $i$ works shift pattern $j$ and 0 otherwise. Let parameters $m$, $n$, $p$ be the number of shift patterns, nurses and grades respectively. Parameter $a_{jk}$ is 1 if shift pattern $j$ covers day/night $k$, 0 otherwise. $q_{is}$ is 1 if nurse $i$ is of grade $s$ or higher, 0 otherwise. Furthermore, $p_{ij}$ is the preference cost of nurse $i$ working shift pattern $j$, and $F(i)$ is the set of feasible shift patterns for nurse $i$. Then we can use the following integer programming formulation from (Dowsland 1998):

$$\text{Min} \sum_{i=1}^{n} \sum_{j \in F(i)}^{m} p_{ij} x_{ij} \tag{1}$$

$$\text{s.t.} \sum_{j \in F(i)} x_{ij} = 1, \forall i = \{1,...,n\}, \tag{2}$$

$$\sum_{j \in F(i)} \sum_{i=1}^{n} q_{is} a_{jk} x_{ij} \geq R_{ks}, \forall k = \{1,...,14\}, s = \{1,...,p\} \tag{3}$$

$$x_{ij} \in \{0,1\}, \forall i, j.$$

The objective function (1) is to minimize total preference cost of all nurses. Constraint (2) ensures that every nurse works exactly one shift pattern from his/her feasible set, and constrain (3) ensures that the demand for nurses is fulfilled for every grade on every day and night. This problem can be regarded as a multiple-choice set-covering problem. The sets are given by the shift pattern vectors and the objective is to minimize the cost of the sets needed to provide sufficient cover for each shift at each grade. The constraints described in (2) enforce the choice of exactly one pattern (set) from the alternatives available for each nurse.

## 3 A BOA for Nurse Scheduling

In the nurse scheduling problem we are tackling, the number of nurses is fixed (approximately 30 depending on data instance), and the goal is to create weekly schedules by assigning each nurse one shift pattern in the most efficient way. The problem can be solved by using a suitable rule, from a rule set that contains a number of available rules, for each nurse's assignment. Thus, a potential solution is represented as a rule string, or a rule sequence connecting all nurses.

We chose this rule-base building approach, as the longer-term aim of our research is to model the explicit learning of a human scheduler. Human schedulers can provide high quality solutions, but the task is tedious and often requires a large amount of time. Typically, they construct schedules based on rules learnt during scheduling. Due to human limitations, these rules are typically simple. Hence, our rules will be relatively simple, too. Nevertheless, human generated schedules are of high quality due to the ability of the scheduler to switch between the rules, based on the state of the current solution. We envisage the proposed BOA to perform this role.

### 3.1 The Construction of a Bayesian Network

Bayesian networks are also called directed graphical models, in which each node corresponds to one variable, and each variable corresponds to one position in the strings representing the solutions. The relationship between two variables is represented by a directed edge between the two corresponding nodes.

Bayesian networks are often used to model multinomial data with both discrete and continuous variables by encoding the relationship between the variables contained in the modeled data, which represents the structure of a problem. Furthermore, they are used to generate new data instances or variables instances with similar properties as those of given data.

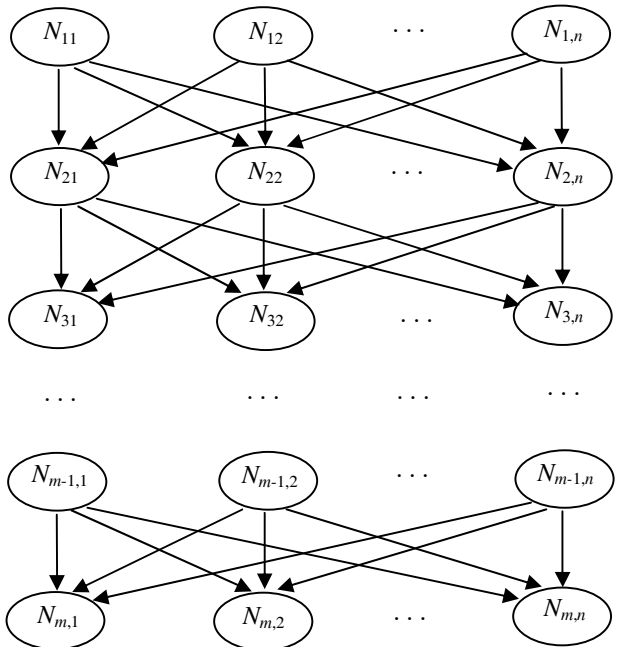

**Fig. 1.** A Bayesian network for nurse scheduling

Figure 1 is the Bayesian network constructed for our nurse scheduling problem, which is a hierarchical and acyclic directed graph representing the solution structure of the problem. The node $N_{ij} (i \in \{1,2,...,m\}; j \in \{1,2,...,n\})$ in the network is a nurse/rule pair which denotes that nurse $i$ is assigned by rule $j$, where $m$ is the number of nurses and n is the number of available rules. The directed edge from node $N_{ij}$ to node $N_{i+1,j'}$ denotes a causal relationship of "$N_{ij}$ causing $N_{i+1,j'}$", i.e. if nurse $i$ is scheduled by rule $j$ then the next nurse will be scheduled by rule $j'$. In this network, a potential solution is shown as a directed path from nurse 1 to nurse m connecting m nodes.

### 3.2 Learning based on the Bayesian Network

In BOAs, the structure of the Bayesian network can be either fixed (Pelikan et al 1999) or variable (Mühlenbein and Mahnig 1999). In our proposed nurse scheduling model, we use a fixed network structure because all variables are fully observed. In essence, our Bayesian network is a fixed nurse-size vector of rules and the goal of learning is to find the variable values of all nodes $N_{ij}$ that maximize the likelihood of the training data containing a number of independent cases.

Thus, the learning in our case amounts to 'counting' based on a multinomial distribution and we use the symbol '#' meaning 'the number of' in the following equations. It calculates the conditional probabilities of each possible value for each node given all possible values of its parent nodes. For example, for node $N_{i+1,j'}$ with a parent node $N_{ij}$, its conditional probability is

$$P(N_{i+1,j'} | N_{ij}) = \frac{P(N_{i+1,j'}, N_{ij})}{P(N_{ij})} = \frac{\#(N_{i+1,j'} = true, N_{ij} = true)}{\#(N_{i+1,j'} = true, N_{ij} = true) + \#(N_{i+1,j'} = false, N_{ij} = true)}. \quad (4)$$

Note that nodes N1j have no parents. In this circumstance, their probabilities are computed as

$$P(N_{1j}) = \frac{\#(N_{1j} = true)}{\#(N_{1j} = true) + \#(N_{1j} = false)} = \frac{\#(N_{1j} = true)}{\#\text{Training sets}}. \quad (5)$$

To help the understanding of how these probabilities are computed, let us consider a simple example of using three rules to schedule three nurses (shown in Figure 2). The scheduling process is repeated 50 times, and each time, rules are randomly used to get a solution, disregarding the feasibility of the solution. The Arabic numeral adjacent to each edge is the total number of times that this edge has been used in the 50 runs. For instance, if a solution is obtained by using rule 2 to schedule nurse 1, rule 3 to schedule nurse 2 and rule 1 to schedule nurse 3, then there exists a path of "N12→N23→N31", and the count of edge "N12→ N23" and edge "N23→N31" are increased by one respectively.

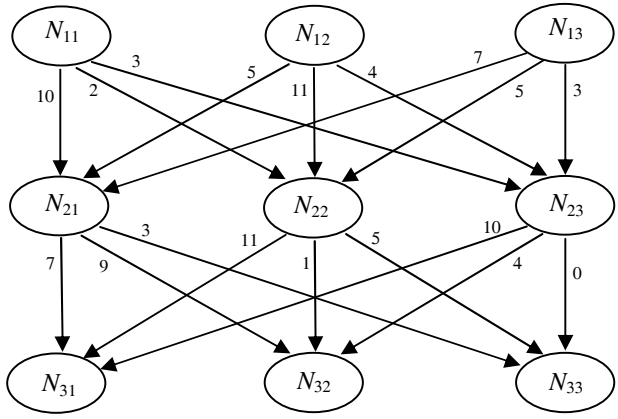

**Fig. 2.** A dummy problem with three nurses and three rules

Therefore, we can calculate the probabilities of each node at different states according to the above count. For the nodes that have no parents, their probabilities are computed as:

$$P(N_{11}) = \frac{10+2+3}{(10+2+3)+(5+11+4)+(7+5+3)} = \frac{15}{50}, \quad P(N_{12}) = \frac{20}{50}, \quad P(N_{13}) = \frac{15}{50}.$$

For all other nodes (with parents), the conditional probabilities are:

$$P(N_{21} | N_{11}) = \frac{10}{10+2+3} = \frac{10}{15}, \quad P(N_{22} | N_{11}) = \frac{2}{15}, \quad P(N_{23} | N_{11}) = \frac{3}{15},$$

$$P(N_{21} | N_{12}) = \frac{5}{5+11+4} = \frac{5}{20}, \quad P(N_{22} | N_{12}) = \frac{10}{20}, \quad P(N_{23} | N_{12}) = \frac{4}{20},$$

$$P(N_{21} | N_{13}) = \frac{7}{7+5+3} = \frac{7}{15}, \quad P(N_{22} | N_{13}) = \frac{5}{15}, \quad P(N_{23} | N_{13}) = \frac{3}{15},$$

$$P(N_{31} | N_{21}) = \frac{7}{7+9+3} = \frac{7}{19}, \quad P(N_{32} | N_{21}) = \frac{9}{19}, \quad P(N_{33} | N_{21}) = \frac{3}{19},$$

$$P(N_{31} | N_{22}) = \frac{11}{11+1+5} = \frac{11}{17}, \quad P(N_{32} | N_{22}) = \frac{1}{17}, \quad P(N_{33} | N_{22}) = \frac{5}{17},$$

$$P(N_{31} | N_{23}) = \frac{10}{10+4+0} = \frac{10}{14}, \quad P(N_{32} | N_{23}) = \frac{4}{14}, \quad P(N_{33} | N_{23}) = \frac{0}{14}.$$

These probability values can be used to generate new rule strings, or new solutions. Since the first rule in a solution has no parents, it will be chosen from nodes N1j according to their probabilities. The next rule will be chosen from nodes Nij according to the probabilities conditioned on the previous nodes. This build-ing process is repeated until the last node Nmj has been chosen, where m is number of the nurses. A path from nurse 1 to nurse m is thus formed, representing a new potential solution. Since all probability values for each nurse are normalized, we suggest the roulette-wheel method as a suitable strategy for rule selection (Goldberg 1989).

For further clarity, consider the following example of scheduling five nurses with two rules (1: random allocation, 2: allocate nurse to low-cost shifts). In the beginning of the search, the probabilities of choosing rule 1 or rule 2 for each nurse is equal, i.e. 50%. After a few iterations, due to the selection pressure and reinforcement learning, we experience two solution pathways: Because pure low-cost or random allocation produces low quality solutions, either rule 1 is used for the first 2-3 nurses and rule 2 on remainder or vice versa. Therefore for this simple example, the Bayesian network learns 'use rule 2 after 2-3 times of using rule 1' or vice versa.

### 3.3 Our BOA Approach

Based on the estimation of conditional probabilities, this section introduces a BOA for nurse scheduling. It uses techniques from the field of modeling data by Bayesian networks to estimate the joint distribution of promising solutions. The nodes, or variables, in the Bayesian network correspond to the individual nurse/rule pairs by which a schedule will be built step by step.

The conditional probability of each variable in the Bayesian network is computed according to an initial set of promising solutions. Subsequently, each new instance for each variable is generated by using the corresponding conditional probabilities, until all variables have been generated. Hence, in our case, a new rule string has been obtained. Another set of rule strings will be generated in this way, some of which will replace previous strings based on fitness selection. If the stopping criterion not reached, the conditional probabilities for all nodes in the Bayesian network are updated again using the current set of promising rule strings. In more detail, the steps of our BOA for nurse scheduling are described as follows:
1. Set t = 0, and generate an initial population P(0) at random;
2. Use roulette-wheel to select a set of promising rule strings S(t) from P(t);
3. Compute the conditional probabilities for the values of each node according to this set of promising solutions;
4. For each nurse's assignment, use the roulette-wheel method to select one rule according to the conditional probabilities of all available nodes for this nurse, thus obtaining a new rule string. A set of new rule strings O(t) will be generated in this way;
5. Create a new population P(t+1) by replacing some rule strings from P(t) with O(t), and set t = t+1;
6. If the termination conditions not met, go to step 2.

### 3.4 Four Heuristic Rules for Solution Building

Using our domain knowledge of nurse scheduling, the BOA can choose from the following four rules. The first two rules are quite simple, while the rest twos are a bit complex. Note that there are many more heuristic rules that could be used to build schedules.

#### 1) 'Random' Rule

This 'Random' rule is used to assign shift patterns to individual nurses in a totally random manner. Its purpose is to introduce randomness into the search thus enlarging the search space, and most importantly to ensure that the proposed algorithm has the ability to escape from local optima. This rule mirrors much of a scheduler's creativeness to come up with different solutions if required.

#### 2) 'k-Cheapest' Rule

The second 'k-Cheapest' rule is purely towards the solution cost, and disregards the feasibility of the solution. For each nurse's assignment, it randomly selects one candidate shift pattern from a set containing the cheapest k shift patterns available to that nurse, in an effort to reduce the solution cost of a schedule as much as possible.

### *3) 'Overall Cover' Rule*

The 'Overall Cover' rule is designed to consider only the feasibility of the schedule. It schedules one nurse at a time in such a way as to cover those days and nights with the most number of uncovered shifts.

This rule constructs solutions as follows. For each shift pattern in a nurse's feasible set, it calculates the total number of uncovered shifts that would be covered if the nurse worked that shift pattern. For instance, assume that a shift pattern covers Monday to Friday nights. Further assume that the current requirements for the nights from Monday to Sunday are as follows: (-4, 0, +1, -3, -1, -2, 0), where a negative number means undercover and a positive overcover. The Monday to Friday shift pattern hence has a cover value of 8 as the sum of undercover is -8. In this example, a Tuesday to Saturday pattern would have a cover value of 6 as the sum of undercover is -6.

For nurses of grade *s*, only the shifts requiring grade s nurses are counted as long as there is a single uncovered shift for this grade. If all these are covered, shifts of the next lower grade are considered and once these are filled those of the next lower grade etc. This operation is necessary as otherwise higher graded nurses might fill lower graded demand, whilst higher graded demand might not be met at all.

Due to the nature of this rule, nurses' preference costs $p_{ij}$ are not taken into account. Therefore, the 'Overall Cover' rule can be summarized as assigning the shift pattern with the largest amount of undercover for the nurse currently being scheduling.

### *4) 'Contribution' Rule*

The fourth 'Contribution' rule is biased towards solution quality but includes some aspects of feasibility. It cycles through all shift patterns of a nurse, assigns each one a score, and chooses the one with the highest score. In case of more than one shift pattern with the best score, the first such shift pattern is chosen. Compared with the third 'Overall Cover' rule, this 'Contribution' rule is more complex because it considers the preferences of the nurses and also tries to look ahead a little.

The score of a shift pattern is calculated as the weighted sum of the nurse's $p_{ij}$ value for that particular shift pattern and its contribution to the cover of all three grades. The latter is measured as a weighted sum of grade one, two and three uncovered shifts that would be covered if the nurse worked this shift pattern, i.e. the reduction in shortfall. Obviously, nurses can only contribute to uncovered demand of their own grade or below.

More precisely, using the same notation as before, the score $S_{ij}$ of shift pattern *j* for nurse *i* is calculated as

$$S_{ij} = w_p(100 - P_{ij}) + \sum_{s=1}^{3} w_s q_{is} (\sum_{k=1}^{14} a_{jk} d_{ks}) \tag{6}$$

where parameter $w_p$ is the weight of the nurse's $p_{ij}$ value for the shift pattern and parameter $w_s$ is the weight of covering an uncovered shift of grade *s*. Variable $a_{jk}$ is 1 if shift pattern *j* covers day *k*, 0 otherwise. Variable $d_{ks}$ is 1 if there are still nurses needed on day *k* of grade *s*, 0 otherwise. Note that $(100-p_{ij})$ must be used in the score, as higher $p_{ij}$ values are worse and the maximum for $p_{ij}$ is 100.

### 3.5 Fitness Function

Our nurse scheduling problem is complicated by the fact that higher qualified nurses can substitute less qualified nurses but not vice versa. Furthermore, the problem has a special day-night structure as most of the nurses are contracted to work either days or nights in one week but not both. These two characteristics make the finding and maintaining of feasible solutions in any heuristic search is extremely difficult. Therefore, a penalty function approach is needed while calculating the fitness of completed solutions. Since the chosen encoding automatically satisfies constraint set (2) of the integer programming formulation, we can use the following formula to calculate the fitness of solutions (the fitter the solution, the lower its fitness value):

$$\text{Min} \sum_{i=1}^{n}\sum_{j=1}^{m} p_{ij} x_{ij} + w_{demand} \sum_{k=1}^{14}\sum_{s=1}^{p} \max\left(\left[R_{ks} - \sum_{i=1}^{n}\sum_{j=1}^{m} q_{is} a_{jk} x_{ij}\right], 0\right). \tag{7}$$

Note that only undercovering is penalized not overcovering, hence the use of the max function. The parameter $w_{demand}$ is the penalty weight used to adjust the penalty that a solution has added to its fitness, and this penalty is proportional to the number of uncovered shifts. For example, consider a solution with an objective function value of 22 that undercovers the Monday day shift by one shift and the Tuesday night shift by two shifts. If the penalty weight was set to 20, the fitness of this solution would be 22 + (1+2)*20 = 82.

## 4 Computational Results

This section describes the computational experiments that are used to test the proposed BOA. For all experiments, 52 real data sets as given to us by the hospital are available. Each data set consists of one week's requirements for all shift and grade combinations and a list of available nurses together with their preference costs pij and qualifications. The data was collected from three wards over a period of several months and covers a range of scheduling situations, e.g. some data instances have very few feasible solutions whilst others have multiple optima.

For all data instances, we used the following set of fixed parameters to implement our experiments. These parameters are based on our experience and intuition and thus not necessarily the best for each instance. We have kept them the same for consistency.

- Stopping criterion: number of generations = 200, or an optimal solution is found;
- Population size = 140;
- The number of solutions kept in each generation = 40;
- For the 'k-Cheapest' rule, k = 5;
- Weight set in formula (6): wp =1, w1 =8, w2 =2 and w3 =1;
- Penalty weight in fitness function (7): wdemand =200;
- Number of runs per data instance = 20;

The BOA was coded in Java 2, and all experiments were run on the same Pentium 4 2.0GHz PC with 512MB RAM under the Windows XP operating system. To test the robustness of the BOA, each data instance was run twenty times by fixing the above parameters and varying the pseudorandom number seed at the beginning. The executing time per run and per data instance varies from half second to half a minute depending on the difficulty of the individual data instance. For example, data instances 27, 29, 31 and 51 are harder to solve than others due to a shortage of nurses in these weeks.

The detailed computational results over these 52 instances are listed in Table 1, in which the last row (headed with 'Av.') contains the mean values of all columns using following notations:

- IP: optimal solutions found with an integer programming software (Dowsland and Thompson, 2000);
- Rd-1: random search, i.e. only the first 'Random' rule is in use;
- Rd-2: random rule-selection, i.e. using four rules but every rule has an equal opportunity to be chosen all the time for all nurses;
- Best: best result out of 20 runs of the BOA;
- Mean: average result of 20 runs of the BOA;
- Fea: number of runs terminating with the best solution being feasible;
- #: number of runs terminating with the best solution being optimal;
- ≤3: number of runs terminating with the best solution being within three cost units of the optimum. The value of three units was chosen as it corresponds to the penalty cost of violating the least important level of requests in the original formulation. Thus, these solutions are still acceptable to the hospital.

**Table** 1. Comparison of results over 52 instances

| Data | IP | RD1 | RD2 | Best | Mean | Fea | # | ≤3 |
|---|---|---|---|---|---|---|---|---|
| 01 | 8 | N/A | 27 | 8 | 8.1 | 20 | 19 | 20 |
| 02 | 49 | N/A | 85 | 56 | 74.5 | 20 | 0 | 0 |
| 03 | 50 | N/A | 97 | 50 | 72.0 | 20 | 2 | 5 |
| 04 | 17 | N/A | 23 | 17 | 17.0 | 20 | 20 | 20 |
| 05 | 11 | N/A | 51 | 11 | 12.4 | 20 | 8 | 16 |

| | | | | | | | |
|---|---|---|---|---|---|---|---|
| 06 | 2 | N/A | 51 | 2 | 2.4 | 20 | 17 | 17 |
| 07 | 11 | N/A | 80 | 14 | 16.8 | 20 | 0 | 3 |
| 08 | 14 | N/A | 62 | 15 | 17.5 | 20 | 0 | 11 |
| 09 | 3 | N/A | 44 | 14 | 17.3 | 20 | 0 | 0 |
| 10 | 2 | N/A | 12 | 2 | 4.9 | 20 | 2 | 10 |
| 11 | 2 | N/A | 12 | 2 | 2.8 | 20 | 2 | 20 |
| 12 | 2 | N/A | 47 | 3 | 7.8 | 20 | 0 | 2 |
| 13 | 2 | N/A | 17 | 3 | 3.6 | 20 | 0 | 20 |
| 14 | 3 | N/A | 102 | 4 | 6.5 | 20 | 0 | 7 |
| 15 | 3 | N/A | 9 | 4 | 5.1 | 20 | 0 | 20 |
| 16 | 37 | N/A | 55 | 38 | 38.8 | 20 | 0 | 20 |
| 17 | 9 | N/A | 146 | 9 | 11.3 | 20 | 4 | 11 |
| 18 | 18 | N/A | 73 | 19 | 20.8 | 20 | 0 | 20 |
| 19 | 1 | N/A | 135 | 10 | 12.0 | 20 | 0 | 0 |
| 20 | 7 | N/A | 53 | 7 | 8.3 | 20 | 5 | 19 |
| 21 | 0 | N/A | 19 | 1 | 1.6 | 20 | 0 | 20 |
| 22 | 25 | N/A | 56 | 26 | 27.5 | 20 | 0 | 15 |
| 23 | 0 | N/A | 119 | 1 | 1.6 | 20 | 0 | 20 |
| 24 | 1 | N/A | 4 | 1 | 1.0 | 20 | 20 | 20 |
| 25 | 0 | N/A | 3 | 0 | 0.2 | 20 | 18 | 20 |
| 26 | 48 | N/A | 222 | 52 | 66.8 | 20 | 0 | 1 |
| 27 | 2 | N/A | 158 | 28 | 35.6 | 20 | 0 | 0 |
| 28 | 63 | N/A | 88 | 65 | 65.5 | 20 | 0 | 3 |
| 29 | 15 | N/A | 31 | 109 | 169.2 | 20 | 0 | 0 |
| 30 | 35 | N/A | 210 | 38 | 83.5 | 20 | 0 | 3 |
| 31 | 62 | N/A | 253 | 159 | 175.0 | 20 | 0 | 0 |
| 32 | 40 | N/A | 102 | 43 | 80.4 | 20 | 0 | 4 |
| 33 | 10 | N/A | 30 | 11 | 15.6 | 20 | 0 | 8 |
| 34 | 38 | N/A | 95 | 41 | 63.8 | 20 | 0 | 2 |
| 35 | 35 | N/A | 118 | 46 | 60.2 | 20 | 0 | 0 |
| 36 | 32 | N/A | 130 | 45 | 47.6 | 20 | 0 | 0 |
| 37 | 5 | N/A | 28 | 7 | 8.2 | 20 | 0 | 7 |
| 38 | 13 | N/A | 130 | 25 | 33.0 | 20 | 0 | 0 |
| 39 | 5 | N/A | 44 | 8 | 10.8 | 20 | 0 | 3 |
| 40 | 7 | N/A | 51 | 8 | 8.8 | 20 | 0 | 10 |
| 41 | 54 | N/A | 87 | 55 | 56.2 | 20 | 0 | 15 |
| 42 | 38 | N/A | 188 | 41 | 73.3 | 20 | 0 | 1 |
| 43 | 22 | N/A | 86 | 23 | 24.2 | 20 | 0 | 13 |
| 44 | 19 | N/A | 70 | 24 | 77.4 | 20 | 0 | 0 |
| 45 | 3 | N/A | 34 | 6 | 8.1 | 20 | 0 | 2 |
| 46 | 3 | N/A | 196 | 7 | 9.4 | 20 | 0 | 0 |
| 47 | 3 | N/A | 11 | 3 | 3.4 | 20 | 13 | 20 |
| 48 | 4 | N/A | 35 | 5 | 5.7 | 20 | 0 | 10 |
| 49 | 27 | N/A | 69 | 30 | 47.8 | 20 | 0 | 2 |
| 50 | 107 | N/A | 162 | 109 | 175.5 | 20 | 0 | 1 |
| 51 | 74 | N/A | 197 | 171 | 173.6 | 20 | 0 | 0 |
| 52 | 58 | N/A | 135 | 67 | 98.6 | 20 | 0 | 0 |
| Av. | 21.1 | N/A | 82.9 | 29.7 | 39.8 | 20 | 2.5 | 8.5 |

According to the computational results in Table 1, one can see that using the 'Random' rule alone does not yield a single feasible solution, as the 'Rd1' column shows. This underlines the difficulty of this problem. In addition, without learning, the results of randomly selecting one of the four rules at each move are much weaker, as the 'Rd2' column shows. Thus, it is not simply enough to use the four rules to build schedules that are acceptable to the hospital for practical use. For the proposed BOA, 38 out of 52 data in-stances are solved to or near to optimality (i.e. within three cost units) and feasible solutions are always found for all instances.

The behavior of an individual run of the BOA is as expected. Figure 3 depicts the BOA search process for data instance 04,. In this figure, the x-axis represents the number of generations and the y-axis represents the solution cost of the best individual in each generation. As shown in Figure 3, the optimal solution, with a cost of 17, is produced at the generation of 57. The actual values may differ among various instances, but the characteristic shapes of the curves are similar for all seeds and data instances.

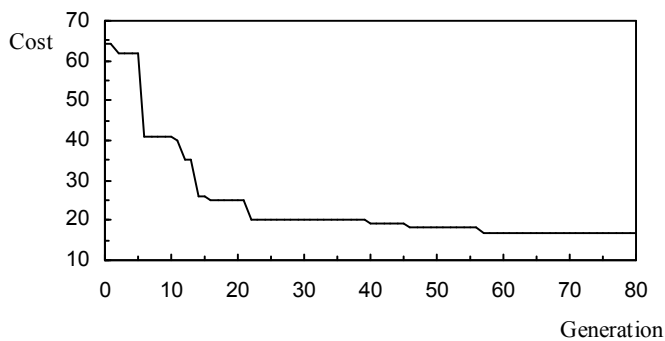

**Fig. 3.** a sample run of the BOA

To understanding the results further, we have developed a graphical interface to visualize the BOA's learning process, instead of simply outputting the final numerical values from a "black box". The software package is developed as a Java applet, which enables the BOA to run real problems online at **http://www.cs.nott.ac.uk/~jpl/BOA/BOA_Applet.html**.

In our graphical interface, the conditional probability of choosing a rule for a specific nurse has been mapped into a 256-gray value and the causal relationship between two nodes (i.e. nurse/rule pairs) is denoted by drawing an edge in this grey value to connect the nodes. For instance, a probability of 50% would give a grey gradient of 128. The darker the edge, the higher the chance this edge (building block) will be used.

Please note that in our online demonstration two further rules can be selected. These rules are still experimental and currently under evaluation. The two additional rules are the 'Highest Cover' rule and the 'Enhanced Contribution' rule. The 'Highest Cover' rule is very similar to the 'Overall Cover' rule described in 3.4, with the difference in their ways of calculating the undercover: a 'Highest Cover' rule is to find shift patterns which cover those days and nights with the highest number of uncover, while a 'Overall Cover' rule is to find the largest total number of uncovered shifts. The second 'Enhanced Contribution' rule is similar to the 'Contribution' rule (described in 3.4), and also uses formula (6) to assign each feasible shift pattern a score. However, this rule uses a different definition of dks: dks is the actual number of required nurses if there are still nurses needed on day k of grade s, 0 otherwise.

Figure 4, 5 and 6 show the graphic results equivalent to the experiments carried out in this paper, i.e. same parameter selection and 4 rules. The pictures show a single run for the '01' data instance which involves the scheduling of 26 nurses. Figure 4 show the beginning, Figure 5 the intermediate and Figure 6 the final stages of the learning process respectively.

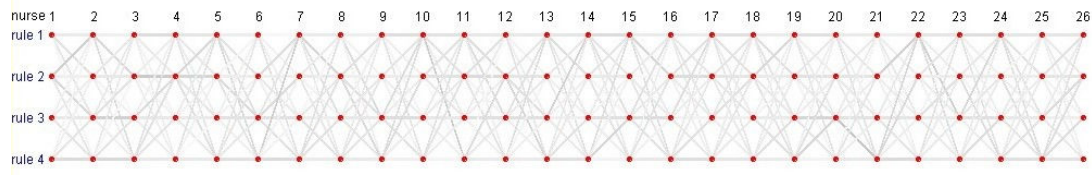

**Fig. 4.** Graphic display of the probability distributions at the initial generation

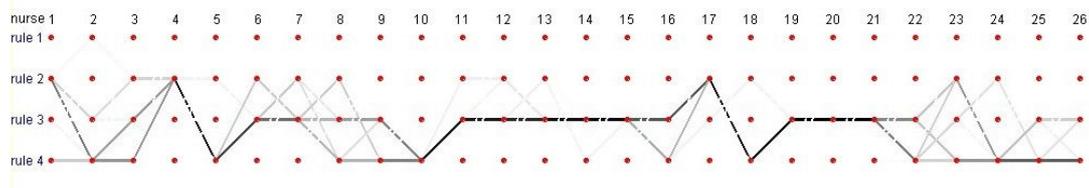

**Fig. 5.** Graphic display of the probability distributions after 50 generations

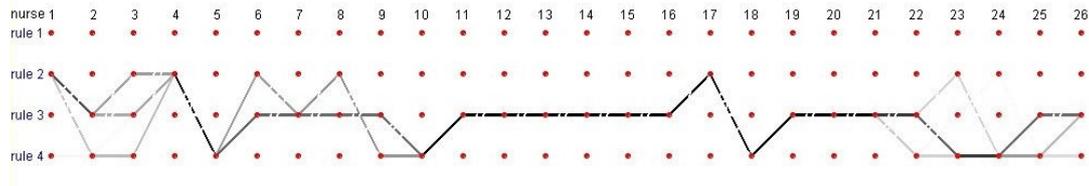

**Fig. 6.** Graphic display of the probability distributions after 100 generations

The graphic results in Figure 4, 5 and 6 are in accordance with our hypothesis. In the early process of the BOA, any edge (i.e. any building block) can be used to explore the solution space as much as possible. Thus, no particular edge stands out in Figure 4. With the BOA in progress, some edges are becoming more remarkable, while some ill-fitting paths become diminishing (Shown in Figure 5). Eventually, the BOA converges to several optimum solutions depicted as clear paths with some common parts, such as the segment of path from node 10 to node 21 (shown in Figure 6). This is particularly true for our problem instances because we know from the hospital that the optimum schedules are often not unique.

Based on the summary results of 20 runs on the same 52 benchmark instances, Table 2 give a comparison of our BOA with some other approaches, which are briefly described as follows:

- LCS: a learning classifier system.
- Basic GA: a GA with standard genetic operators;
- Adaptive GA: the same as the basic GA, but the algorithm also tries to self-learn good parameters during the runtime;
- Multi-population GA: the same as the adaptive GA, but it also features competing sub-populations;
- Hill-climbing GA; the same as the multi-population GA, but it also includes a local search in the form of a hill-climber around the current best solution;
- Indirect GA: a novel GA-based approach that first maps the constrained solution space into an unconstrained space, then optimizes within that new space and eventually translates solutions back into the original space. Up to four different rules and hill-climbers are used in this algorithm.

**Table 2.** Summary results of various approaches

| Algorithm | Reference | Best | Mean | Feasibility | Runtime |
|---|---|---|---|---|---|
| BOA | Li and Aickelin 2003 | 29.7 | 39.8 | 100% | 23.0 secs |
| LCS | Li and Aickelin 2004 | 35.5 | 47.7 | 100% | 44.5 secs |
| Optimal IP | Dowsland and Thompson 2000 | 21.4 | 21.4 | 100% | >24 hrs |
| Basic GA | Aickelin and White 2004 | 79.8 | 88.6 | 33% | 18.9 secs |
| Adaptive GA | Aickelin and White 2004 | 65.0 | 71.4 | 45% | 23.4 secs |
| Multi-population GA | Aickelin and White 2004 | 37.1 | 59.8 | 75% | 13.4 secs |
| Hill-climbing GA | Aickelin and White 2004 | 23.2 | 44.7 | 89% | 14.9 secs |
| Indirect GA | Aickelin and Dowsland 2003 | 22.0 | 25.5 | 95% | 11.9 secs |

Let us discuss the summary results in Table 2. Compared with the optimal results of the IP approach which can take more than 24 hours runtime for each instances, the BOA's results are still more expensive on average but they are achieved within half minute. Compared with other meta-heuristic approaches which are all executed quickly, our BOA performs best in terms of feasibility. In terms of solution quality, in general our BOA performs better than the LCS, the basic GA, the adaptive GA and the multi-population GA, and performs slightly worse than the hill-climbing GA and the indirect GA. However, it is worth mentioning that the hill-climbing GA includes the features of multiple populations and elaborate local search which are not available in the proposed BOA, and the indirect GA uses the best nurses' ordering together with a local search to produce the final solution while our BOA only uses the ordering given in the original data sets throughout the search.

## 5  Conclusions

A BOA is presented for the nurse scheduling problem. Unlike most existing rule-based approaches, the proposed BOA has the ability to build schedules by using flexible, rather than fixed rules. Experimental results from real-world nurse scheduling problems have demonstrated the strength of the approach.

Although we have presented this work in terms of nurse scheduling, it is suggested that the main idea of the BOA could be applied to many other scheduling problems where the schedules will be built systematically according to specific rules. It is also hoped that this research will give some preliminary answers about how to include human-like learning into scheduling algorithms and thus may be of interest to practitioners and researchers in areas of scheduling and evolutionary computation.

## Acknowledgements

The work was funded by the UK Government's major funding agency, the Engineering and Physical Sciences Research Council (EPSRC), under grants GR/R92899/02.